\documentclass[12pt]{article}

\usepackage[round]{natbib}
\usepackage{multirow,graphicx}
\RequirePackage{amsthm,amsmath}
\RequirePackage[colorlinks,citecolor=blue,
urlcolor=blue,backref]{hyperref}
\usepackage{color}

\usepackage{enumerate,booktabs,xcolor,verbatim,soul}
\newcommand{\equald} {{\buildrel d \over =}}

\def\bM {\mathbf{M}}

\def\bT {\mathbf{T}}
\def\bx {\mathbf{x}}
\def\bX {\mathbf{X}}

\def\bX {\mathbf{X}}
\def\bY {\mathbf{Y}}
\def\bW {\mathbf{W}}

\def\bV {\mathbf{V}}
\def\bG {\mathbf{G}}
\def\vkl{\nu_{kl}}

\def\bZ {\mathbf{Z}}

\DeclareMathOperator*{\argmin}{\arg\,\min}

\providecommand{\keywords}[1]{\textbf{\textit{Keywords and phrases: }} #1}
\providecommand{\mscclass}[1]{\textbf{\textit{AMS 2010 subject classifications: }} #1}

\def\uN{{\rm N }}

\def\T{{\rm T}}
\def\M{{\rm M}}
\setstcolor{red}

\usepackage[pagewise]{lineno}
\begin{document}
\title{Bayesian quantile additive regression trees}
\author{ Bereket P. Kindo \and 
Hao Wang \and Timothy  Hanson \and Edsel A. Pe\~na}
\date{\vspace{-5ex}}
\maketitle

\begin{abstract}
Ensemble of regression trees have become popular statistical tools for the estimation of conditional mean given a set of predictors. However, quantile regression trees and their ensembles have not yet garnered much attention despite the increasing popularity of the linear quantile regression model. This work proposes a Bayesian quantile additive regression trees model that shows very good predictive performance illustrated using simulation studies and real data applications. Further extension to tackle binary classification problems is also considered. 

\end{abstract}

\let\thefootnote\relax\footnotetext{
}
\let\thefootnote\relax\footnotetext{\mscclass{Primary 62F15; Secondary 62H12 }	
}

\label{chap:bayes_q_art}
\section{Introduction}
\label{sec:intro-quantile}

Quantile regression gives a comprehensive picture of the relationship between a response variable and a set of predictors. It is particularly appealing when the
 inferential interest lies in the probabilistic properties of extreme observations 
 conditional on a set of predictors. Such objectives arise in various disciplines: in environmental sciences, \cite{friederichs2007statistical} study the probabilistic properties of extreme precipitation events, while \cite{pedersen2015predictable} model the tail distribution of stock and bond returns. In an epidemiological study, \cite{burgette2011exploratory} use penalized quantile regression to explore covariates that affect the lower tail of  the distribution of birth weight of babies. When the distribution of the dependent variable is skewed, the desire for robustness to extreme observations
  makes quantile regression a preferred approach. Examples include the study of tourist expense patterns in \cite{marrocu2015micro} and wage distribution in  \cite{buchinsky1995quantile}.  
 
Extensive work in the theory and application of linear quantile regression can be found in 
\cite{koenker1978regression,koenker1994confidence, buchinsky1998recent,tsai2012relationship,cole1992smoothing}.  Suppose we have a data set $\left(y_i, \mathbf{x}_i \right) \text{ for } i = 1, \ldots, n$, where
$y_i \in \Re$ and $\mathbf{x}_i \in  \Re^d$ denote the
observed response and predictors for the $i^{\text{th}}$ observation, respectively. Analogous to the use of the mean function $E(y|\bx)$ used in least squares regression to explain the relationship 
between the response and predictors, quantile regression uses the $\tau^{\text{th}}$ quantile function $Q\left(y |\bx, \tau \right)$, where $\tau \in \left(0,1\right)$. The $\tau^{\text{th}}$ quantile  of a random variable $Y$ with distribution $F$ is defined as $Q(\tau) = 
\inf{\left\{y: F(y) \geq \tau \right\}}$, where $F\left( \cdot \right)$ denotes the cumulative distribution function. Thus, for a given quantile value $\tau$, quantile regression seeks to estimate $Q(\bx, \tau) = 
\inf{\left\{y: F\left(y \vert \bx \right) \geq \tau \right\}}$ .
The linear quantile regression problem in particular  is described as the minimization problem 
\begin{equation}
\label{eqn:quantile_reg_minimization}
	\hat{\beta}_{\tau} = \argmin_{\beta} \displaystyle \sum_{i=1}^{n}{\rho_{\tau}\left(y_i - \bx_i^{\texttt{T}}\beta \right)},
\end{equation}
where $\rho_{\tau}(\omega) = \omega \left( \tau - I\left\{\omega < 0 \right\}\right)$ is usually termed as the ``check loss'' function. The error distribution is left largely unspecified except that its $\tau^{\text{th}}$ quantile equals zero. The work in \cite{koenker1978regression} spearheaded the use of quantile regression as a robust alternative to mean regression. More recently, $l_1$ regularized quantile regression with simultaneous variable selection and parameter estimation is studied in \cite{zou2008composite,belloni2011_l1}.

\begin{figure}
\centering
\includegraphics[width=3.5in, scale = 0.5]{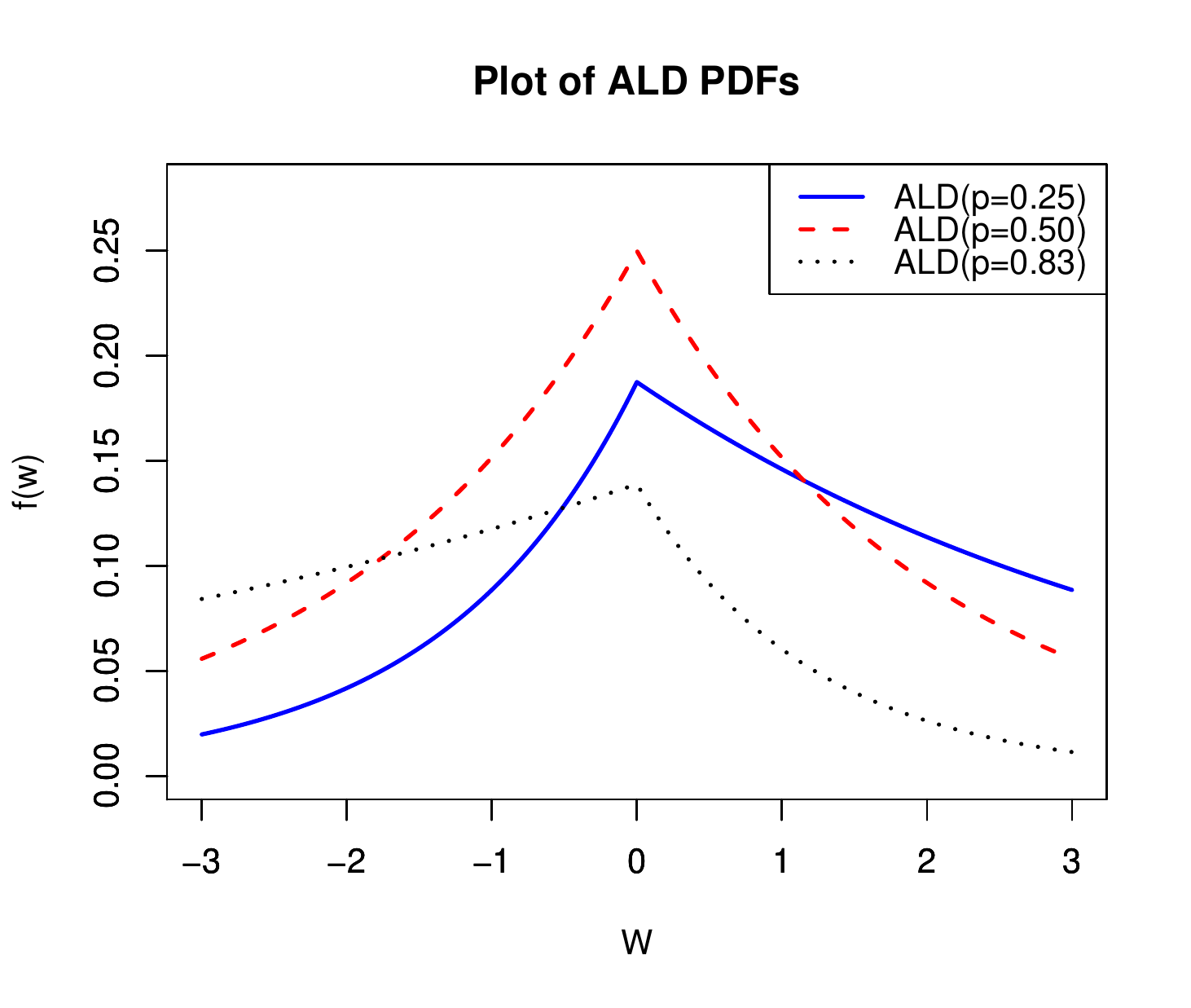}
\caption{Asymmetric Laplace distribution with $\mu = 0$ and values of $\tau \in \left\{0.25, 0.50, 0.83 \right\}$.  }
\label{fig:aldpdfs}
\end{figure}

An alternative, yet equivalent, formulation of \eqref{eqn:quantile_reg_minimization} assumes that the random errors follow the asymmetric Laplace distribution \citep{yu2001bayesian,kozumi2011gibbs,sriram2013posterior}. If a random variable $Y$ follows an asymmetric Laplace distribution $\text{ALD}(y; \tau, \mu)$ with location parameter $\mu \in \Re$, its density function is given by 
\begin{equation}
\label{eqn:ald_density}
f_{\tau} \left( y; \mu \right)  =  \tau \left(1 - \tau \right) \exp \left\{-  \rho_{\tau} \left( y - \mu \right) \right\},
\end{equation}
where $\tau \in (0,1), \rho_{\tau}(\omega) = \omega \left( \tau - I\left\{\omega < 0 \right\}\right)$ for $\omega \in \Re$. A special case of \eqref{eqn:ald_density} with $\tau = 0.5$ is the Laplace double exponential distribution.  Figure \ref{fig:aldpdfs} shows the plots of the probability density functions of asymmetric Laplace distributions for fixed location parameter $\mu = 0$, and values of $\tau \in \left\{0.25, 0.50, 0.83 \right\}$. The expectation and variance of $Y \sim \text{ALD}(\tau, \mu=0)$ are 
\begin{equation}
\text{E} \left( Y \right) = \frac{1 - 2\tau}{\tau \left( 1-\tau \right)} \text{ and } \text{Var}\left(Y \right) = \frac{1-2\tau+2\tau^2}{\tau^2 \left(1 - \tau \right)^2},
\end{equation}
while its characteristic function is $ \psi_Y \left(t \right) = \left[\frac{1}{2} \vartheta_2^2 t^2 - \vartheta_1 t i + 1 \right]^{-1}$,
where $\vartheta_1 = \frac{1 - 2\tau}{\tau(1-\tau)}$ and $\vartheta_2^2 = \frac{2}{\tau(1-\tau)}$. 

Some Bayesian approaches to the quantile regression problem in general and median regression in particular have been considered in \cite{yu2001bayesian,
dunson2005approximate,
taddy2012bayesian,
hanson2002modeling,
kozumi2011gibbs,
kottas2001bayesian,
reich2010flexible} either by assuming asymmetric Laplace,  Dirichlet process mixtures, Polya trees, or Gaussian mixture approximations as the distribution of the random error term. In particular, \cite{kozumi2011gibbs} outline a Gibbs sampler for Bayesian quantile regression based on a mixture representation of the asymmetric Laplace distribution. With the intention of utilizing their approach, we paraphrase their finding which they show using the equality of characteristic functions. 
If the random variables $V$ and $Z$ which follow the standard exponential and Gaussian distributions respectively are mutually independent, then $W = \vartheta_1 V + \vartheta_2 \sqrt{V} Z$ is equal in distribution to the asymmetric Laplace distribution $\text{ALD}(\tau, \mu=0)$, where $\vartheta_1 = \frac{1-2\tau}{\tau \left(1-\tau \right)}$ and $\vartheta_2^2 = \frac{2}{\tau\left(1 - \tau \right)}$. Such representation allows a formulation of an efficient algorithm to estimate regression quantiles in a Bayesian framework that involves simulations from the Gaussian and Generalized Inverse Gaussian distributions.

In comparison to least squares regression trees, quantile regression trees or their ensembles have not yet garnered much attention. However, sporadic works in the literature exist including the single tree quantile regression model  of \cite{chaudhuri2002nonparametric}
and the quantile regression forests model in \cite{meinshausen2006quantile} which extends on the idea of random forests \citep{breiman2001random}. In  quantile regression forests model of \cite{meinshausen2006quantile}, all of the observations that lie in a regression tree terminal node are used for estimation while a summary statistic (typically the average) of the observations in a terminal node are used by random forests. At the core of the quantile regression forests is the empirical estimation of the conditional cumulative density function $F\left(y \vert \bx \right) = P\left(Y \leq y \vert \bx \right)$ so that $\hat{Q}(\bx, \tau) = \inf{\left\{y: \hat{F}\left(y \vert \bx \right) \geq \tau \right\}},$ where $\hat{F}$ is an estimator of $F$.

Bayesian regression trees and their ensembles are shown to have enhanced predictive performance in the framework of least squares regression, and binary and multiclass classification \citep{CGM98, CGM10,Abu_Nimeh2008, ZhaEtAl2010, pratola2014parallel,kapelner2013bartmachine,kindo2016mpbart} 
. In particular, BART - Bayesian additive regression trees \citep{CGM10} estimates the conditional mean of a response given a set of predictors by using a sum of regression trees model
\begin{equation}
\label{eqn:bart_model}
y  = \sum_{j=1}^{n_T}{g \left(\bx; \T_j, \M_j \right)} + \epsilon, \text{ where } \epsilon \sim \uN \left(0, \sigma^2\right).
\end{equation}
BART is specified through priors on the regression trees via a ``tree generating stochastic process'' that favors shallow trees and prior specifications on terminal node parameters that strategically shrink the influence of individual trees. BART has been utilized in many applications with great predictive performance \citep{Abu_Nimeh2008, ZhaEtAl2010,wu2010novel,
he2009profiling,liu2015ensemble}. In this article we explore the utility of ensemble of Bayesian regression trees to garner a comprehensive view of the dependence between a response and predictors. Thus, we propose a fully Bayesian framework for construction of quantile regression trees and their  ensembles to complement the linear Bayesian quantile regression of \cite{kozumi2011gibbs,yu2001bayesian} and quantile regression forests of \cite{meinshausen2006quantile}. We note that, at the time of this writing, we are not aware of a Bayesian counterpart in the literature to the frequentist quantile regression tree.

The remaining parts of this article are outlined as follows. Section \ref{sec:bayes_qart} sets the framework for Bayesian quantile additive regression trees including the prior specifications on all the parameters of the model and posterior computations. Section \ref{sec:data_analysis} delves into the implementation of the model with simulation studies and real data applications. Section \ref{sec:binary_extension} extends Bayesian quantile additive trees to tackle binary classification problems along with a simulation study and real data application. Section \ref{sec:conclusion} provides concluding remarks.

\section{Bayesian quantile additive regression trees}
\label{sec:bayes_qart}

In this section we outline the model specifications for Bayesian quantile additive regression trees. Specifically, let the observable data be $\left(y_i, \mathbf{x}_i \right) \text{ for } i = 1, \ldots, n$, where
$y_i \in \Re$ and $\mathbf{x}_i \in  \Re^d$ denoting the response and predictors for the $i^{\text{th}}$ observation. Consider the model
\begin{equation} 
\label{eqn:sum_of_quantile_tree_model}
 \begin{aligned}
 y_i & =  \bG \left(\bx_i; \bT, \bM \right) + \vartheta_1 \nu_{i} + \vartheta_2 \phi^{\frac{1}{2}} \sqrt{\nu_{i}} z_i, \\
\bG \left(\bx_i; \bT, \bM \right)  & =  \sum_{j=1}^{n_{\texttt{T}}}  g \left(\bx_i; \T_j, \M_j \right) \\
 p \left( \nu_{i} \vert \phi \right) & =  \frac{1}{\phi} \exp\left\{-\frac{\nu_{i}}{\phi} \right\},\\
 p \left( z_i \right) & = \frac{1}{\sqrt{2 \pi}} \exp\left\{-\frac{1}{2} z_i^2 \right\},
\end{aligned}
\end{equation}
where $\T_j$ and $\M_j$ are the $j^{\text{th}}$ tree in the sum and its associated terminal node parameters, and $\left( \bT, \bM \right) = \left\{ \left( \T_j, \M_j \right); j=1, \ldots, n_{\texttt{T}} \right\}$. Note that $\vartheta_1 \nu_{i} + \vartheta_2 \phi^{\frac{1}{2}} \sqrt{\nu_{i}} z_i = \phi \left[ \vartheta_1 \tilde{\nu}_{i} + \vartheta_2  \sqrt{\tilde{\nu}_{i}} z_i \right],$ where $\tilde{\nu}_i \sim Exp \left( 1 \right)$ and the quantity in the square brackets is the mixture representation of the asymmetric Laplace distribution.

\subsection{Prior specifications}
\label{subsec:sum_tree_prior}
We assume that the priors on any two distinct 
trees in the sum are independent and the 
prior on $\phi$ is independent of the tree 
priors. That is, $\left(\T_j, \M_j\right) \perp \left(\T_{j'}, \M_{j'} \right) \text{ for } j \neq j',$ and $\left(\bM, \bT \right) \perp \phi.$ Further assuming that 
given a tree, say the $j^{\text{th}}$ tree $\T_j$, the priors on its $m_{j}$ terminal node 
parameters are independent enables writing 
the prior distribution on $\left(\bT,\bM,\phi \right)$ as 
\begin{equation}
\label{eqn:sum_tree_priors}
\begin{aligned}
p \left(\bT,\bM,\phi \right) & = \left[\prod_{j=1}^{n_{\texttt{T}}}{ p\left(\T_j, \M_j \right)} \right] p \left(\phi \right) \\
	  					  & = \left[\prod_{j=1}^{n_{\texttt{T}}}{ \left[ p \left( \T_j \right) {p \left( \M_j \vert \T_j \right)} \right]} \right] p \left(\phi \right)\\
	  					  & = \left[\prod_{j=1}^{n_{\texttt{T}}}{ \left[ p \left( \T_j \right) \prod_{k=1}^{m_j} {p\left(\mu_{jk} \vert \T_j \right)} \right]} \right] p \left(\phi \right),
\end{aligned}
\end{equation}
where $n_{\texttt{T}}$ is the number of trees in the sum and $m_j$ is the number of terminal nodes of tree $\T_j$ (i.e., $\M_j = \left(\mu_{j1}, \ldots, \mu_{j{m_j}} \right)$).

The prior $p \left( \T_j \right)$ is specified through a ``tree generating stochastic process'' of \cite{CGM98}. This process is governed by tree splitting rule that creates non-overlapping partitions of the predictor space by selecting a splitting variable followed by a splitting value given the selected variable. Once a terminal node is randomly selected for use in binary partitioning of the predictor space, a splitting variable is randomly chosen followed by a random selection of a value in the range of the selected predictor with condition that no empty partition is created. Furthermore, the probability that a terminal node $\eta$ with depth $d_{\eta}$ (number of ancestor nodes) splits is given by 
\begin{equation}
\label{eqn:prob_grow}
p_{\texttt{SPLIT}}(\eta) = \begin{cases} 
      1 & \text{ if } d_{\eta} =  0 \\
      \frac{\psi_1}{(1 + d_{\eta})^ {\psi_2}}, & \text{ if } d_{\eta} >  0, 
   \end{cases} 
\end{equation}
where $\psi_1 \in (0,1), \psi_2 \in [0,\infty).$ The splitting probability in \eqref{eqn:prob_grow}, and the choice of $\psi_1$ and $\psi_2$ play a crucial role of regulating the influence of individual trees in the sum. For example, higher values of $\psi_2$ and lower values of $\psi_1$ result in shallow trees in general. 

Given a tree $\T_j$, the prior on the terminal node parameters  is a Gaussian distribution $\mu_{jk} \vert \T_j \sim \uN \left(\mu_0 , \sigma^2_0  \right)$ for $k = 1, \ldots, m_j$. In the model representation given in  \eqref{eqn:sum_of_quantile_tree_model}, the overall contribution of the prior distributions of the terminal node parameters on $\text{E}\left(y \vert \mathbf{x} \right)$ and $\text{Var}\left(y \vert \mathbf{x} \right)$ are 
$n_{\texttt{T}} \mu_0$ and $n_{\texttt{T}} \sigma^2_0.$ The hyper-parameters $\mu_0$ and $\sigma^2_0$ are 
selected so that the overall effect induced 
by the prior distributions is in the interval $\left(\min(y), \max(y)\right)$ with high 
probability. A convenient 
aspect of the quantile function is its 
invariance to a monotone transformation. In 
particular, we use the transformation $\tilde{y} =  h(y) = \frac{y - \min(y)}{\left(\max(y) - \min(y) \right)} - 0.5$ for 
which we have $Q(y,\tau) = h^{-1}(Q( \tilde{y}, \tau))$. 
Taking $\tilde{y}$ as the dependent variable in \eqref{eqn:sum_of_quantile_tree_model} 
along with priors  $\mu_{jk} \vert \T_j \sim \uN \left(\mu_0 = 0, \sigma^2_0 = \frac{1}{ 2\kappa \sqrt{n_{\texttt{T}}}} \right)$, we 
ensure that the transformed response is in 
the interval $\left(-0.5,0.5 \right).$ This 
choice of the hyper-parameters also ensures 
that the effect of the prior distributions on the terminal 
nodes places high probability to the same 
interval. We find that a value of $\kappa$ 
between 2 and 3 gives reasonable results. Note that the larger the number of trees in 
the sum, the smaller the prior 
variance placed on the terminal node 
parameters effectively shrinking the influence of individual trees to zero.   Finally, the prior on $\phi$ is specified as an Inverse-Gamma distribution $\phi \sim IG \left( \frac{\alpha}{2} , \frac{\beta}{2} \right)$.

\subsection{Posterior updating scheme}
\label{subsec:posterior_scheme}
The posterior updating scheme cycles through the following three posterior draws: a draw from
\begin{equation}
\label{eqn:lat_draw}
p \left(\bV \vert \bT, \bM , \bY, \phi \right)
\end{equation}
followed by 
consecutive updates of the $j^{\text{th}}$ tree and its terminal node parameters for $j = 1, \ldots, n_{\texttt{T}}$ accomplished by a draw from 
\begin{equation}
\label{eqn:tree_draws}
p \left\{ \left( \T_j, \M_j \right) \vert \bM_{(-j)}, \bT_{(-j)}, \phi, \bX, \bY \right\},
\end{equation} 
with $\left( \bT_{(-j)}, \bM_{(-j)} \right)$ denoting all the trees and their terminal node parameters in the sum excluding the $j^{th}$ tree;  and finally a draw from 
\begin{equation}
\label{eqn:phi_draw}
p \left\{ \phi \vert \bM, \bT, \bX, \bY\right\},
\end{equation}
where $\bV = \left(\nu_1,\ldots, \nu_n \right)^{\texttt{T}}$, $\bY = \left(y_1,\ldots, y_n \right)^{\texttt{T}}$, and $\bX = \left(\bx_1,\ldots, \bx_n \right)^{\texttt{T}}$. The posterior draw in \eqref{eqn:lat_draw} is $n$ sequential samples from the Generalized Inverse Gaussian distribution 
\begin{equation}
p \left( \nu_{i} \vert \bT, \bM, \phi, y_i, \bx_i \right) \propto \nu_{i}^{-\frac{1}{2}} \exp\left\{- \frac{1}{2} \left[\delta_{1i} \nu_{i}^{-1} + \delta_{2i} \nu_{i} \right] \right\},
\end{equation}
where $\delta_{1i} = \frac{\left( y_i - \bG \left(\bx_i; \bT, \bM \right) \right)^2}{\vartheta_2^2 \phi} $ and $\delta_2 =  \frac{2 \vartheta_2^2 + \vartheta_1^2}{\vartheta_2^2 \phi}$. To describe the draw in \eqref{eqn:tree_draws}, we re-write \eqref{eqn:sum_of_quantile_tree_model} as 
\begin{equation}
\label{eqn:resid_model}
\omega_i \equiv y_i  - \sum_{l \ne j} g \left(\bx_i; \T_l, \M_l \right) - \vartheta_1 \nu_{i}=  g \left(\bx_i; \T_j, \M_j \right)  + \phi^{\frac{1}{2}} \vartheta_2 \sqrt{\nu_{i}} z_i
\end{equation}
so that $\omega_i \vert \bx_i, \nu_i, \bT_{(-j)}, \bM_{(-j)}, \phi \sim \uN \left(g \left(\bx_i; \T_j, \M_j \right), \phi \vartheta_2^2 \nu_i \right).$  A Metropolis-Hastings algorithm is utilized to update the tree $\T_j$ with $\bW = \left(\omega_1, \ldots, \omega_n \right)^{\texttt{T}}$ considered a residual psuedo-response variable. A similar Bayesian ``back-fitting'' algorithm is implemented in \cite{CGM10,kindo2016mpbart}.

For ease of explanation of the Metropolis-Hasting algorithm, we pursue a slight modification of notation as follows. Suppose that $\bW_k = \left(\omega_{k1}, \ldots, \omega_{kn_k} \right)^{\texttt{T}}$ is a vector of residuals that lie in the $k^{\text{th}}$ terminal node of the regression tree $\T_j$  which has $m_j$ terminal nodes, and that $\bX_k  = \left(\bx_{k1}, \ldots, \bx_{kn_k} \right)^{\texttt{T}}$ denotes the corresponding set of predictors. Likewise, $\mathbf{V}_k = \left(\nu_{k1}, \ldots,\nu_{kn_{k}} \right)^{\texttt{T}}$ and $\bZ_k = \left(z_{k1}, \ldots,z_{kn_{k}} \right)^{\texttt{T}}$ denote the components of the mixture representation of asymmetric Laplace error term corresponding to the observations in the $k^{\text{th}}$ terminal node. With this notation, we write $\bW = \left(\bW_1, \ldots, \bW_{m_j} \right)^{\texttt{T}},$
$\bX  = \left( \bX_1, \ldots, \bX_{m_j} \right)^{\texttt{T}},$ and $\bV = \left(\bV_1, \ldots, \bV_{m_j} \right)^{\texttt{T}},$ where $n = n_1 + \ldots + n_{m_j}$.  Similar notation is used in \cite{CGM98}.
We can then write the likelihood function of the single residual tree in \eqref{eqn:resid_model} as
\begin{equation}
f(\bW \vert \bX, \bV, \phi, \T_j, \M_j) = \prod_{k=1}^{m_j}{f(\bW_k \vert \bX_k, \bV_k, \phi, \T_j, \M_j)},
\end{equation}
where
\begin{equation}
\begin{aligned}
& f(\bW_k \vert \bX_k, \bV_k, \phi, \T_j, \M_j) = f(\bW_k \vert \mu_{jk} , \bV_k, \phi) \\
 & = \left[ \frac{1}{\sqrt{2 \pi}\vartheta_2 \phi^{\frac{1}{2} }} \right]^{n_k} \prod_{l=1}^{n_{k}} {\vkl^{- \frac{1}{2}}} \exp{\left\{-\frac{1}{2 \vartheta_2^2 \phi} \sum_{l=1}^{n_k} { \frac{(\omega_{kl} - \mu_{jk} )^2}{\vkl} }  \right\}}.
\end{aligned}
\end{equation}
With the prior specification $\mu_{jk} \sim \uN \left(\mu_0 = 0, \sigma^2_0 = \frac{1}{2 \kappa \sqrt{n_{\texttt{T}}}} \right),$ we have  
\begin{equation}
\begin{aligned}
& \int f(\bW_k, \M_j \vert \bX_k, \T_j, \bV_k, \phi) d \M_j \\
& = \int f(\bW_k \vert \bX_k, \T_j, \bV_k, \phi) p \left(\mu_{jk} \right) d \mu_{jk}\\
& = \left[ \frac{1}{\sqrt{2 \pi}\vartheta_2 \phi^{\frac{1}{2}}} \right]^{n_k} 
 \left[\prod_{l=1}^{n_{k}} {\vkl^{- \frac{1}{2}}} \right] \exp{\left\{ - \frac{1}{2 \vartheta_2^2 \phi}  \sum_{l=1}^{n_k} {\omega_{kl}^2 \vkl^{-1} }  \right\}} \times \\
& \quad \sqrt{\frac{ \vartheta_2^2 \phi}{\vartheta_2^2 \phi + \sigma_0^2 \sum_{l=1}^{n_k}{\vkl^{-1}}}} \exp \left\{\ \frac{\sigma_0^2 \left[ \sum_{l=1}^{n_k} {\omega_{kl} \vkl^{-1} } \right]^2}{2 \vartheta_2^2 \phi \left[ \vartheta_2^2 \phi + \sigma_0^2 \sum_{l=1}^{n_k}{\vkl^{-1}}\right]} \right\}.
\end{aligned}
\end{equation}
To draw from $p \left\{ \left( \T_j, \M_j \right) \vert \bM_{(-j)}, \bT_{(-j)}, \phi, \bX, \bY \right\}$, we first obtain a tree $\T_j^*$ as a candidate update to $\T_j$ accepted with a probability 
\begin{equation}
                \label{eqn:accept_prob2}
                \min \left\{1, \frac{q(\T^*_j, \T_j) p(\bW \vert \bX, \bV, \T^*_j, \phi ) p(\T^*_j)}{q(\T_j, \T^*_j) p(\bW \vert \bX, \bV, \T_j, \phi) p(\T_j)}\right\}.
\end{equation}
The transition kernel $q\left(\cdot, \cdot \right)$ assigns probabilities of 0.25, 0.25, 0.40 and 0.10 to the moves GROW, PRUNE, SWAP, and CHANGE respectively. The GROW move randomly selects a terminal node and proposes a binary split with probability of \eqref{eqn:prob_grow} while its reverse counterpart PRUNE move randomly selects and collapses a pair of terminal node parameters originating from the same parent node. The CHANGE move randomly selects a non-terminal node and changes the splitting variable and value. It affects terminal nodes that are  descendants of the node where CHANGE move is applied. However, this move does not change the number of terminal and non-terminal nodes. The SWAP move interchanges the splitting rule of a parent and child non-terminal nodes. 

For illustrative purposes, we elaborate on the calculation of the ratio 
\begin{equation}
\label{eqn:ratio}
\frac{p(\bW \vert \bX, \bV, \T^*_j, \phi )}{p(\bW \vert \bX, \bV, \T_j, \phi )},
\end{equation}
which is a component of \eqref{eqn:accept_prob2}. For the fittingly named GROW move, when a terminal node with $n_p$ observation splits to left and right nodes of size $n_l$ and $n_r$ (the subscripts $p$, $l$ and $r$ denoting ``parent'', and ``left'' and ``right'' child nodes), \eqref{eqn:ratio} 
simplifies through cancellations since a GROW move only affects the terminal node that is being split. That is, 
\begin{equation}
\label{eqn:ratio2}
\begin{aligned}
\frac{p(\bW \vert \bX, \bV, \T^*_j, \phi )}{p(\bW \vert \bX, \bV, \T_j, \phi )} & = \frac{p(\bW_l \vert \bX_l, \bV_l, \T^*_j, \phi ) p(\bW_r \vert \bX_r, \bV_r, \T^*_j, \phi )}{p(\bW_p \vert \bX_p, \bV_p, \T_j, \phi )}
\end{aligned}
\end{equation}
which equals
\begin{equation*}
\begin{aligned}
&  \sqrt{\frac{ \vartheta_2^2 \phi \left( \vartheta_2^2 \phi + \sigma_0^2 B_p\right)}{ \left( \vartheta_2^2 \phi + \sigma_0^2 B_r \right) \left( \vartheta_2^2 \phi + \sigma_0^2 B_l \right) } }  \times \\ 
& \exp \left\{\ \frac{\sigma_0^2 }{2 \vartheta_2^2 \phi }  \left(\frac{A_r^2}{\vartheta_2^2 \phi + \sigma_0^2 B_r} + \frac{A_l^2}{\vartheta_2^2 \phi + \sigma_0^2 B_l} - \frac{A_p^2}{\vartheta_2^2 \phi + \sigma_0^2 B_p}\right) \right\} ,
\end{aligned}
  \end{equation*} 
where $B_k = \sum_{l=1}^{n_l}{\vkl^{-1}}$ and $A_k = \sum_{l=1}^{n_k} {\omega_{kl} \vkl^{-1} }$ whose dependence on $\bV_k$ and $\bW_k$ is suppressed for conciseness.
Given an updated tree $\T_j$, its terminal node parameters $\M_j = \left(\mu_{jk}; k=1, \ldots, m_j \right)$ are updated by drawing from $p \left( \mu_{jk} \vert \T_j, \bV, \phi, \bW, \bX \right)$ which upto a proportionality constant is  given by
\begin{equation}
 \exp{\left\{- \frac{1}{2} \left( \frac{\vartheta_2^2 \phi + \sigma_0^2 \sum_{l=1}^{n_k}{\vkl^{-1}}}{\vartheta_2^2 \sigma^2_0 \phi} \right) \left[ \mu_{jk} - \frac{\sigma^2_0 \sum_{l=1}^{n_k} {\omega_{kl} \vkl^{-1} } }{\vartheta_2^2 \phi + \sigma_0^2 \sum_{l=1}^{n_k}{\vkl^{-1}}} \right]^2 \right\}},
\end{equation}
indicating a sample from a Gaussian distribution. 

In order to update the scale parameter $\phi$, we revert to the original notation of the quantile sum of trees in \eqref{eqn:sum_of_quantile_tree_model}, then draw from Inverse-Gamma distribution
\begin{equation}
p \left(\phi \vert \bM, \bT, \bY, \bX, \bV \right) \propto \phi^{-\frac{n}{2} - \frac{\alpha}{2} -1} \exp \left\{- \frac{1}{\phi} \left[ \frac{\beta}{2} + \sum_{i=1}^{n}\frac{\left(y_i - \mathbf{G}(\bx_i) - \vartheta_1 \nu_i \right)^2}{2 \vartheta_2^2 \nu_i } \right] \right\}.
\end{equation}

\section{Data analysis}
\label{sec:data_analysis}
\subsection{Simulation study}
\label{subsec:simulation}
\begin{figure}
\centering
\includegraphics[width=3.75in, height = 3.0in, scale = 0.25]{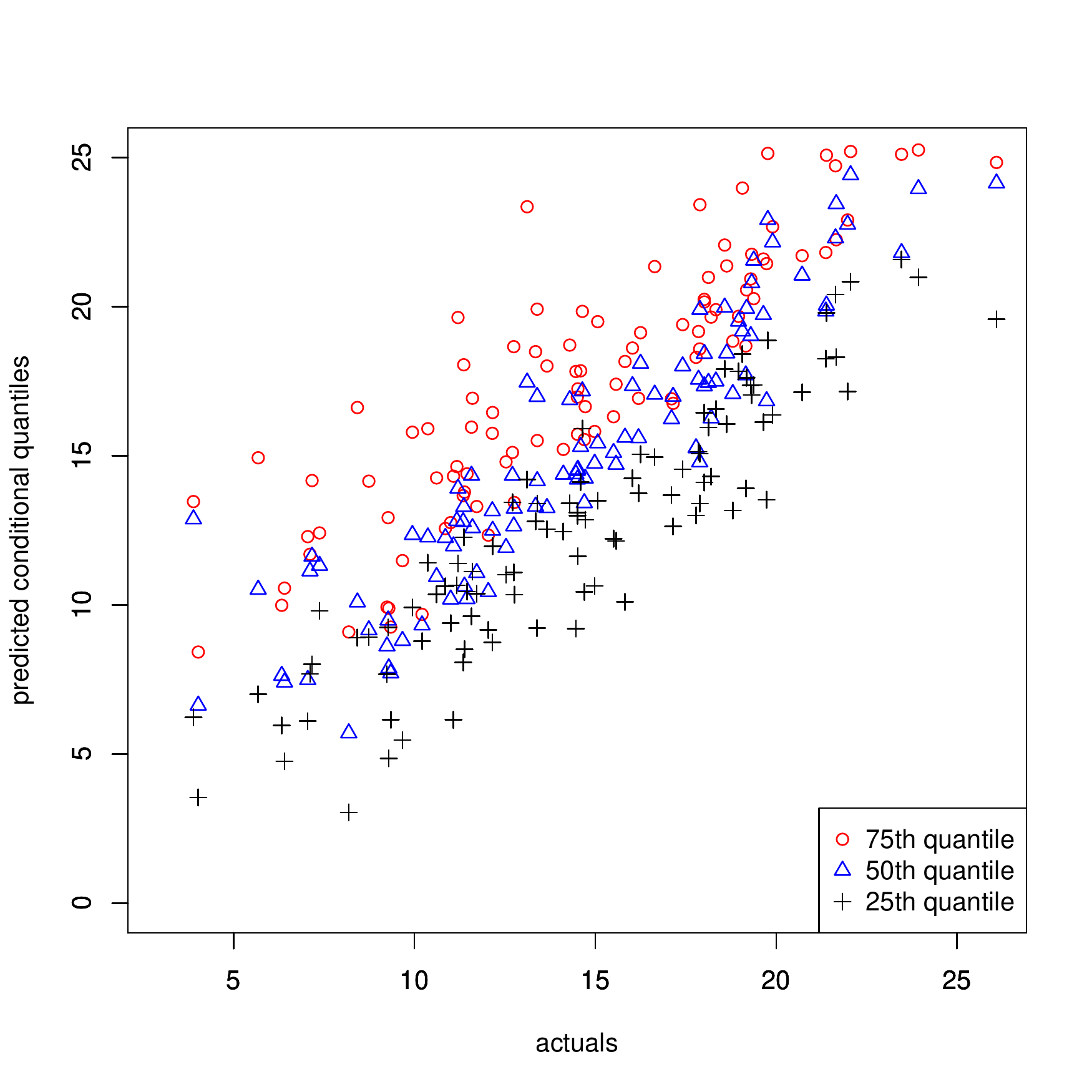}
\caption{Predicted conditional quantiles against the actual response for the first simulation study. }
\label{fig:actual_vs_predicted}
\end{figure}

In this subsection, two simulation studies are conducted. The first uses the function $f: \Re^{10} \to \Re$ given by $f(\mathbf{x}) =   10 \sin(\pi x_1 x_2) + 20(x_3 - 0.5)^2+10 x_4+5 x_5 + 0(x_6+x_7+x_8+x_9+x_{10}),$ where $x_j \sim \text{Unif} (0,1) \text{ for } j = 1, \ldots, 10$. This benchmark data generating function is used in \cite{friedman1991multivariate, CGM10,gramacy2012cases} among others. The response variable is simulated as $y 
= f \left( \mathbf{x} \right) + \epsilon,$ where $\epsilon \, \equald \, \pi \epsilon_1 + \left(1 - \pi\right) \epsilon_2$, $\pi \sim \text{Bern} (0.8)$, $\epsilon_1 \sim \uN (0,1)$ and $\epsilon_2 \sim 
\uN (1,4)$. Note that it includes non-linear, linear, interaction effects as well as predictors that do not affect the response variable. The model evaluation metric used is the mean weighted absolute deviation given by 
\begin{equation}
\label{eqn:wmad}
\text{MWAD} = \frac{1}{n} \sum_{i = 1}^{n} \rho_{ \tau
}(\hat{y}_i - y_i),
\end{equation} 
where  $\rho_{ \tau}(\omega)= \omega \left(  \tau - I\left\{\omega < 0 \right\}\right)$ is the 
``check'' loss function, $\hat{y}_i$ is the estimated conditional $ \tau^{\text{th}}$ quantile and 
$y_i$ is the actual response value of the $i$th observation in the evaluation data set. Twenty replications of training and test data sets of $100$ observations each are simulated and test data set performance comparisons for Bayesian quantile additive regression trees (BayesQArt), Bayesian 
quantile regression with adaptive Lasso regularization (BayesQR.AL) in \cite{alhamzawi2012bayesian,li2010bayesian}, linear regression quantiles with adaptive Lasso regularization (QReg.AL) in \cite{zou2008composite} and quantile random forest (QRF) in \cite{meinshausen2006quantile} are reported. Our proposed method shows very good predictive performance with lower mean weighted absolute deviation  than the competing procedures in estimating the $25^{\text{th}}$, $50^{\text{th}}$ and $75^{\text{th}}$ conditional quantiles as displayed in Table \ref{tab:sim1_set1}, underscoring its robustness to the presence of intricate relationships between predictors and the dependent variable. Figure \ref{fig:actual_vs_predicted} displays the predicted conditional quantiles against the actual response values. 

\begin{table}
\centering
\begin{tabular}{llll}    
\toprule
	& $\tau = 0.25$ &  $\tau = 0.50$ & $\tau =  0.75$ \\
\midrule
BayesQArt & 0.7190	 (0.0243) & 0.9236 (0.0228) &  0.6795 (0.0170)\\
\midrule
QRF  & 0.9215 (0.0228)  & 1.1430 (0.0243)   &1.0123 (0.0171)  \\
\midrule
BayesQR.AL & 0.8577 (0.0065)  & 1.0274 (0.0069)  & 0.8298 (0.0083) \\
\midrule
QReg.AL & 0.8395 (0.0194) & 1.0132 (0.0211) & 0.8130 (0.0151) \\
\bottomrule
\end{tabular}
	\caption{First quantile regression simulation study results: test data average mean weighted absolute deviations (MWADs) \eqref{eqn:wmad} and standard errors in parentheses over 20 replications. }\label{tab:sim1_set1}
\end{table}

In the second simulation study, a data set with 30 predictors of which 10 do not impact the response in any form is generated. The data generating scheme is based on the heteroskedastic error model in \cite{he1997quantile}
\begin{equation}
y = \bx^{\texttt{T} } \beta  + \left( \bx^{\texttt{T}}  \gamma \right) \epsilon, 
\end{equation} 
where $\epsilon \sim \uN \left(0,1 \right)$, $\bx \in \Re^{30}$, 
$\beta =
\begin{pmatrix}
 1_{20 \times 1} \\
 0_{10 \times 1}
 \end{pmatrix},$
$\gamma =
\begin{pmatrix}
 1_{5 \times 1} \\
 0_{25 \times 1}
 \end{pmatrix},$ and $1_{m \times 1}$ and $0_{m \times 1}$ denoting column vectors of ones and zeros of size $m$.  Each component of $\bx$ is generated independently from $\text{Unif} \left(0,1\right)$. The results of this simulation study for estimation of $25^{\text{th}}$, $50^{\text{th}}$ and $75^{\text{th}}$ conditional quantiles is reported in Table \ref{tab:sim_2_results} on twenty replications of training and test data sets  of 100 observations each. For the estimation of the $50^{\text{th}}$ and $75^{\text{th}}$ conditional quantiles, the linear models have better performance than our method or quantile random forest. This is expected given the underlying data generating process assumes a linear relationship between the predictors and the dependent variable. Our method performs well showing better results than quantile random forests for the estimation of the $50^{\text{th}}$ and $75^{\text{th}}$ conditional quantiles. 
 
\begin{table}
\centering
\begin{tabular}{llll}    
\toprule
	& $\tau = 0.25$ &  $\tau = 0.50$ &  $\tau = 0.75$   \\
\midrule
BayesQArt & 0.4864 (0.0137)  & 0.5367 (0.0087)  & 0.3619 (0.0045)   \\
\midrule
QRF  & 0.4601 (0.0179) & 0.6327 (0.0166) & 0.5143 (0.0067) \\
\midrule
BayesQR.AL & 0.7601 (0.0042) &0.5083 (0.0028) & 0.2549 (0.0014) \\
\midrule
QReg.AL & 0.7624 (0.0042) &0.5082 (0.0028)&0.2541 (0.0014)  	\\
\bottomrule
\end{tabular}
	\caption{Second quantile regression simulation study results:  test data average mean weighted absolute deviations (MWADs) \eqref{eqn:wmad} and standard errors in parentheses based on 20 replications. }\label{tab:sim_2_results}
\end{table}

\subsection{Real data examples}
The first data used for illustrating Bayesian quantile additive regression trees is the airquality data from the \textit{R} package \textit{datasets}. This data set records the ozone levels (in parts per billion) in New York from May  to September 1973. The predictors used are a measure of solar radiation level, wind speed, maximum daily temperature, and month and day of measurement.  We estimate the $25^{\text{th}}$, $50^{\text{th}}$ and $75^{\text{th}}$ conditional quantile ozone level using competing statistical procedures in Section \ref{subsec:simulation} and Bayesian quantile additive regression trees. After removing observations with missing records, we split the data into five nearly equal partitions. Table \ref{tab:ozone_result} reports the mean weighted absolute deviations and standard errors.    
\begin{table}
\centering
\begin{tabular}{llll}    
\toprule
	& $\tau = 0.25$ &  $\tau = 0.50$ & $\tau =  0.75$ \\
\midrule
BayesQArt & 4.7855 (0.4967)  & 6.7465 (0.5402)   & 6.0362 (0.7636)   \\
\midrule
QRF  & 4.4835 (0.7453)  & 6.4043 (0.7731) & 5.7270 (0.6917) \\
\midrule
BayesQR.AL & 5.9865 (0.4938) & 7.6987 (0.6479)  & 7.5316 (0.7150) \\
\midrule
QReg.AL & 6.0649 (0.5171) & 8.7799 (0.3412) &  8.0962 (0.3104) \\
\bottomrule
\end{tabular}
	\caption{Ozone data set: test data average mean weighted absolute deviations (MWADs) \eqref{eqn:wmad} and standard errors in parentheses based on 5 consecutive splits of the data as they appear in the \textit{R} package \textit{datasets}. }\label{tab:ozone_result}
\end{table}

The second real data set considered is an auto insurance data consisting of 2,812 auto insurance policyholders with 56 predictors along with an aggregate  paid claim amount. This data set is available in the \textit{R} package \textit{HDtweedie} \citep{qian2015tweedie}. Examples of the predictors are driver's age, driver's income, use of vehicle (commercial or not), vehicle type (either of 6 categories), and driver's gender. The response variable is the aggregate claim amount and it is skewed with substantial policyholders having zero claims. When the claim amounts are non-zero, larger claim amounts tend to be reported. 

Insurers are often interested in understanding the distribution of claim amounts conditional on a set of policyholder and policy characteristics with added emphasis on higher quantiles. Neither the existence of claims nor the amount if a claim occurs is known at the time of the policy purchase. Hence, insurers use estimates of future claims to appropriately price the insurance product and also to set aside sufficient amount of monetary reserves to pay future claims. Thus, we estimate the $90^{\text{th}}$ and $95^{\text{th}}$ conditional quantiles by splitting the data set into 10 nearly equal partitions each time using nine-tenth of the data for training and the remaining for testing Bayesian quantile additive regression trees and the statistical procedures in Section \ref{subsec:simulation}. Table \ref{tab:auto_result} displays the predictive performances of each procedure and our method performs very well. Note that we are intentionally using the regularized versions of the procedures Bayesian linear quantile regression and the classical quantile regression since variable selection is a component of these procedures.  

\begin{table}
\centering
\begin{tabular}{lll}    
\toprule
	& $\tau = 0.90$ &  $\tau = 0.95$  \\
\midrule
BayesQArt & 1.4487 (0.0690)   &  1.0440 (0.0571)  \\
\midrule
QRF  & 1.4862 (0.0676)   & 1.0656 (0.0522)  \\
\midrule
BayesQR.AL &  1.4508 (0.0681) & 1.0483 (0.0602) \\
\midrule
QReg.AL & 1.4542 (0.0671) & 1.0559 (0.0603)	\\
\bottomrule
\end{tabular}
	\caption{Auto insurance claims data set: test data average mean weighted absolute deviations (MWADs) \eqref{eqn:wmad} and standard errors in parentheses based on 10 splits. }\label{tab:auto_result}
\end{table}

\section{Binary classification extension}
\label{sec:binary_extension}
In this section, we extend Bayesian quantile additive regression trees to tackle binary classification problems.  \cite{kordas2006smoothed,benoit2012binary, Benoit2013} among others consider the binary classification problem in a quantile regression framework. Suppose $y_i \in \left\{0, 1\right\}$ and $\bx_i \in \Re^d$ are the binary response and the predictors for the $i^{\text{th}}$ observation. Suppose also that there is an unobserved latent variable $\tilde{y}_i$ for $i=1, \ldots, n$ such that $y_i = 1$  if $\tilde{y}_i > 0$ and $y_i = 0$ otherwise.
The goal of the classification problem is to obtain an estimate $\hat{P}\left(y_i = 1 \mid \bx_i \right)$ for $P\left( y_i = 1 \mid \bx_i \right)$ which we obtain via the hierarchical model 
\begin{equation} 
\label{eqn:binary__model}
 \begin{aligned}
y_i \vert \tilde{y}_i, \nu_i, \bG, \bT, \bM & \sim \text{Bern} \left( \text{P} \left(\tilde{y}_i > 0 \mid  \nu_i, \bG, \bT, \bM  \right)\right) 
\\
\tilde{y}_i \mid  \nu_i, \bG, \bT, \bM & \sim \uN \left(\bG \left(\bx_i; \bT, \bM \right) + \vartheta_1 \nu_{i} , \vartheta_2^2 \phi \nu_{i}\right) 
\\
\bG \left(\bx_i; \bT, \bM \right)  & =  \sum_{j=1}^{n_{\texttt{T}}}  g \left(\bx_i; \T_j, \M_j \right) 
\\
\nu_{i} \mid \phi  & \sim  \text{Exp} \left( \phi \right),
\\
z_i & \sim \uN \left(0, 1 \right),
 \end{aligned}
\end{equation}
where $\bG \left(\bx_i; \bT, \bM \right)$ is a sum of regression trees. The prior specifications for $\bT$, $\bM$ are as specified in \eqref{eqn:sum_of_quantile_tree_model}.  The posterior computation cycles through the following MCMC steps. Sequential draws from truncated normal distributions to sample from the latent variable $\tilde{y}_i \textrm{ for } i=1, \ldots, n$
\begin{equation}
\label{eqn:bin_latent}
\begin{aligned}
\tilde{y}_i \mid y_i, \bx_i, \nu_i, \bT, \bM,\phi \sim &\uN \left(\bG \left(\bx_i; \bT, \bM \right) + \vartheta_1 \nu_{i} , \vartheta_2^2 \phi \nu_{i}\right) \text{I}\left(y_i = 1, \tilde{y}_i \geq 0 \right) \\
& + \uN \left(\bG \left(\bx_i; \bT, \bM \right) + \vartheta_1 \nu_{i} , \vartheta_2^2 \phi \nu_{i}\right) \text{I}\left(y_i = 0, \tilde{y}_i < 0 \right),
\end{aligned}
\end{equation}
followed by draws from $\bV, \left(\bT, \bM \right),$ and $\phi$ as described in Section
 \ref{subsec:posterior_scheme} with $\tilde{\bY} = \left(\tilde{y}_1, \ldots, \tilde{y}_n \right)^{\texttt{T}}$ considered as the response vector of the Bayesian quantile additive regression model.  

\subsection{Binary classification simulation study}
\label{subsec:binary_sim}
\begin{figure}
\centering
\includegraphics[width=3.5in, height = 3.0in, scale = 0.25]{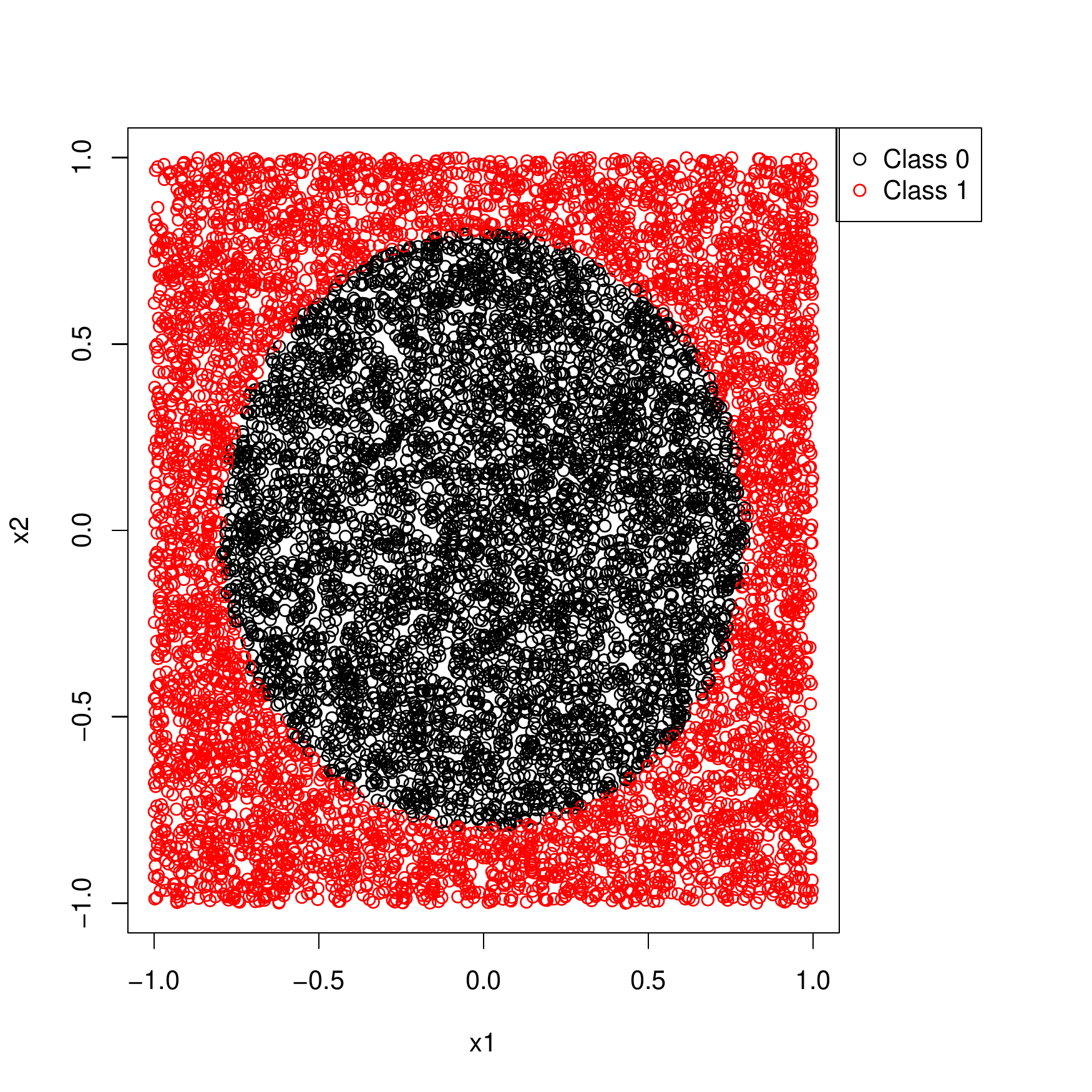}
\caption{circle in square - two dimensional illustration of class boundaries of the binary classification simulation study for Bayesian quantile additive regression trees.  }
\label{fig:scatter_circle}
\end{figure}

We simulate a binary classification data set with ten predictors using the data generating scheme known as ``cicle'' from the \textit{R} package \textit{mlbench} \citep{leisch2010mlbench} which has often been considered as a benchmark classification data set \citep{chung2015accurate,
ishwaran2015effect,rudnicki2015all}. Suppose $\bx \in \left[-1,1\right]^d,$ with $x_j \sim \text{Unif} \left(-1,1 \right), \ j=1, \ldots, d,$ 
where we take $d=10$. The goal of this classification problem is to identify if the coordinate $\left(x_1, \ldots, x_{d} \right)$
 in a $d$ dimensional hypercube with edges at all sign permutations of the coordinates $\left\{ \pm 1, \ldots, \pm 1 \right\}$ lies outside of a hypersphere which lies inside the hypercube. That is, $y = 1$ if $\sum_{j=1}^{d}{x_j^2} > r^2$, otherwise $y = 0$. The radius of the hypersphere, $r,$ is chosen so that there is nearly equal representation between the two classes. The class boundaries are non-linear making it an interesting classification problem (see Figure \ref{fig:scatter_circle} which shows the class boundary for the two dimensional case). 

We simulate training and test data sets of size 100 each and report the averages of classification error rate and area under the ROC curve over twenty replications for binary Bayesian quantile additive regression trees (BayesQArt), binary Bayesian linear quantile regression (BayesQR) and random forests (RF). Note that the random forest procedure used for classification in this section is one described in \cite{breiman2001random} and not the quantile random forest in \cite{meinshausen2006quantile}. For an evaluation data set with $m$ observations, classification error rate is computed as 
$$\text{Error Rate} = \frac{1}{m}\sum_{i=1}^{m}{y_i \neq \hat{y}_i}.$$  
\begin{table}
\centering
\begin{tabular}{lcc}
\toprule
Procedure & Error Rate & AUC  \\
\midrule
BayesQArt	& 0.2185 (0.0100)	  & 0.8297 (0.0088) \\
\midrule
RF 	& 0.2600 (0.0112) 	& 0.6363 (0.0130)      \\
\midrule
BayesQR.AL 	&0.5500 (0.0133)	& 0.4684 (0.0078) \\
\bottomrule
\end{tabular}
	\caption{Simulation study for binary classification: test 
	data averages of test data classification error rate, area under the ROC curve (AUC),
	and their standard errors in parentheses based on 20 replications. }\label{tab:sim1_binary_results}
\end{table}

\subsection{Real data for binary classification}
\label{subsec:class_realdata}

We consider a binary classification real data example in which the number of predictors is much larger than the sample size to illustrate the predictive performance of the binary extension of Bayesian quantile additive regression trees. The goal for this data set is to classify whether a patient has cancer (ovarian or prostate cancer) based on 10,000 predictors of which a portion is mass-spectra data and the other portion consisting of unimportant predictors. This data set is obtained from a data set named ``arcene'' at the UCI machine learning repository \citep{BacheLichman2013}. The training and validation data sets combined contain 200 patient samples. Additional details on this data set are in \cite{guyon2007competitive, guyon2008feature}. We split the data into five nearly equal partitions and report the averages of test data classification error rate and area under the ROC curve for binary Bayesian quantile additive regression trees (BayesQArt) and random forests (RF). Results in Table \ref{tab:cancer_class_binary_results} show that our proposed method handles regression problems in which the number of predictors is much larger than the number samples in the training data while exhibiting very good predictive performance. For this example, we only used the moves GROW and PRUNE to reduce to the computational cost. An average computing time of 5.42 minutes is recorded using a 64-bit Windows personal computer with the specifications: \textit{Intel Core i5-2320, 3.0GHz and 6.0GB} installed memory.

\begin{table}
\centering
\begin{tabular}{lcc}
\toprule
Procedure & Error Rate & AUC  \\
\midrule
BayesQArt	&0.1650 (0.0170) &  0.8712 (0.0165)\\
\midrule
RF 	&0.1800 (0.0094) &  0.8184 (0.0110) \\
\bottomrule
\end{tabular}
	\caption{Cancer classification results: 
	test data averages of classification error rate, area under the ROC curve (AUC),
	and their standard errors. }\label{tab:cancer_class_binary_results}
\end{table}


\section{Conclusion}
\label{sec:conclusion}

This article proposed a Bayesian sum of regression trees model for estimating conditional quantiles. The  asymmetric Laplace distribution likelihood is employed with its mixture representation enabling tractable posterior computation of the regression trees in the sum and their terminal node parameters.  

Simulation studies with data generating schemes that included linear, non-linear, interaction effects as well as unimportant predictors illustrated that Bayesian quantile additive regression trees has very good predictive performance. Real data applications dealing with insurance claims and ozone level prediction demonstrated that the proposed method complements existing powerful statistical procedures. 

We also successfully extended and tested the proposed procedure to tackle binary classification problems. The proposed method exhibited very good out-of-sample classification accuracy in a simulation study characterized by a non-linear class boundary and  cancer classification example in which the number of predictors is about fifty times as much as the number of samples in the training data. The source code for the implementation of our proposed method, and the selected tuning parameters for the simulation studies and real data applications are at https://github.com/bpkindo/bayesqart.

\bibliography{bib_bq_art}

\begin{thebibliography}{49}
\providecommand{\natexlab}[1]{#1}
\providecommand{\url}[1]{\texttt{#1}}
\expandafter\ifx\csname urlstyle\endcsname\relax
  \providecommand{\doi}[1]{doi: #1}\else
  \providecommand{\doi}{doi: \begingroup \urlstyle{rm}\Url}\fi

\bibitem[Abu-Nimeh et~al.(2007)Abu-Nimeh, Nappa, Wang, and Nair]{Abu_Nimeh2008}
Saeed Abu-Nimeh, Dario Nappa, Xinlei Wang, and Suku Nair.
\newblock A comparison of machine learning techniques for phishing detection.
\newblock In \emph{Proceedings of the Anti-phishing Working Groups 2nd Annual
  eCrime Researchers Summit}, eCrime '07, pages 60--69. ACM, 2007.
\newblock ISBN 978-1-59593-939-5.

\bibitem[Alhamzawi et~al.(2012)Alhamzawi, Yu, and
  Benoit]{alhamzawi2012bayesian}
Rahim Alhamzawi, Keming Yu, and Dries~F Benoit.
\newblock {Bayes}ian adaptive {Lasso} quantile regression.
\newblock \emph{Statistical Modelling}, 12\penalty0 (3):\penalty0 279--297,
  2012.

\bibitem[Bache and Lichman(2013)]{BacheLichman2013}
K.~Bache and M.~Lichman.
\newblock {UCI} machine learning repository, 2013.
\newblock URL \url{http://archive.ics.uci.edu/ml}.

\bibitem[Belloni and Chernozhukov(2011)]{belloni2011_l1}
Alexandre Belloni and Victor Chernozhukov.
\newblock l1-penalized quantile regression in high-dimensional sparse models.
\newblock \emph{The Annals of Statistics}, 39\penalty0 (1):\penalty0 82--130,
  2011.

\bibitem[Benoit and Van~den Poel(2012)]{benoit2012binary}
Dries~F Benoit and Dirk Van~den Poel.
\newblock Binary quantile regression: a {Bayes}ian approach based on the
  asymmetric {Lapl}ace distribution.
\newblock \emph{Journal of Applied Econometrics}, 27\penalty0 (7):\penalty0
  1174--1188, 2012.

\bibitem[Benoit et~al.(2013)Benoit, Alhamzawi, and Yu]{Benoit2013}
Dries~F. Benoit, Rahim Alhamzawi, and Keming Yu.
\newblock {Bayes}ian {Lasso} binary quantile regression.
\newblock \emph{Computational Statistics}, 28\penalty0 (6):\penalty0
  2861--2873, 2013.

\bibitem[Breiman(2001)]{breiman2001random}
Leo Breiman.
\newblock Random forests.
\newblock \emph{Machine Learning}, 45\penalty0 (1):\penalty0 5--32, 2001.

\bibitem[Buchinsky(1995)]{buchinsky1995quantile}
Moshe Buchinsky.
\newblock Quantile regression, box-cox transformation model, and the {US} wage
  structure, 1963--1987.
\newblock \emph{Journal of Econometrics}, 65\penalty0 (1):\penalty0 109--154,
  1995.

\bibitem[Buchinsky(1998)]{buchinsky1998recent}
Moshe Buchinsky.
\newblock Recent advances in quantile regression models: A practical guideline
  for empirical research.
\newblock \emph{Journal of Human Resources}, 33\penalty0 (1), 1998.

\bibitem[Burgette et~al.(2011)Burgette, Reiter, and
  Miranda]{burgette2011exploratory}
Lane~F Burgette, Jerome~P Reiter, and Marie~Lynn Miranda.
\newblock Exploratory quantile regression with many covariates: an application
  to adverse birth outcomes.
\newblock \emph{Epidemiology}, 22\penalty0 (6):\penalty0 859--866, 2011.

\bibitem[Chaudhuri and Loh(2002)]{chaudhuri2002nonparametric}
Probal Chaudhuri and Wei-Yin Loh.
\newblock Nonparametric estimation of conditional quantiles using quantile
  regression trees.
\newblock \emph{Bernoulli}, 8\penalty0 (5):\penalty0 561--576, 2002.

\bibitem[Chipman et~al.(1998)Chipman, George, and McCulloch]{CGM98}
Hugh~A Chipman, Edward~I George, and Robert~E McCulloch.
\newblock {Bayes}ian {CART} model search.
\newblock \emph{Journal of the American Statistical Association}, 93\penalty0
  (443):\penalty0 935--948, 1998.

\bibitem[Chipman et~al.(2010)Chipman, George, and McCulloch]{CGM10}
Hugh~A Chipman, Edward~I George, and Robert~E McCulloch.
\newblock {BART}: {Bayes}ian additive regression trees.
\newblock \emph{The Annals of Applied Statistics}, 4\penalty0 (1):\penalty0
  266--298, 2010.

\bibitem[Chung and Kim(2015)]{chung2015accurate}
Dongjun Chung and Hyunjoong Kim.
\newblock Accurate ensemble pruning with {PL-b}agging.
\newblock \emph{Computational Statistics \& Data Analysis}, 83:\penalty0 1--13,
  2015.

\bibitem[Cole and Green(1992)]{cole1992smoothing}
Timothy~J Cole and Pamela~J Green.
\newblock Smoothing reference centile curves: the {LMS} method and penalized
  likelihood.
\newblock \emph{Statistics in Medicine}, 11\penalty0 (10):\penalty0 1305--1319,
  1992.

\bibitem[Dunson and Taylor(2005)]{dunson2005approximate}
David~B Dunson and Jack~A Taylor.
\newblock Approximate {Bayes}ian inference for quantiles.
\newblock \emph{Nonparametric Statistics}, 17\penalty0 (3):\penalty0 385--400,
  2005.

\bibitem[Friederichs and Hense(2007)]{friederichs2007statistical}
P~Friederichs and A~Hense.
\newblock Statistical downscaling of extreme precipitation events using
  censored quantile regression.
\newblock \emph{Monthly Weather Review}, 135\penalty0 (6), 2007.

\bibitem[Friedman(1991)]{friedman1991multivariate}
Jerome~H Friedman.
\newblock Multivariate adaptive regression splines.
\newblock \emph{The Annals of Statistics}, pages 1--67, 1991.

\bibitem[Gramacy and Lee(2012)]{gramacy2012cases}
Robert~B Gramacy and Herbert~KH Lee.
\newblock Cases for the nugget in modeling computer experiments.
\newblock \emph{Statistics and Computing}, 22\penalty0 (3):\penalty0 713--722,
  2012.

\bibitem[Guyon et~al.(2007)Guyon, Li, Mader, Pletscher, Schneider, and
  Uhr]{guyon2007competitive}
Isabelle Guyon, Jiwen Li, Theodor Mader, Patrick~A Pletscher, Georg Schneider,
  and Markus Uhr.
\newblock Competitive baseline methods set new standards for the {NIPS} 2003
  feature selection benchmark.
\newblock \emph{Pattern Recognition Letters}, 28\penalty0 (12):\penalty0
  1438--1444, 2007.

\bibitem[Guyon et~al.(2008)Guyon, Gunn, Nikravesh, and Zadeh]{guyon2008feature}
Isabelle Guyon, Steve Gunn, Masoud Nikravesh, and Lofti~A Zadeh.
\newblock \emph{Feature extraction: foundations and applications}, volume 207.
\newblock Springer, 2008.

\bibitem[Hanson and Johnson(2002)]{hanson2002modeling}
Timothy Hanson and Wesley~O Johnson.
\newblock Modeling regression error with a mixture of {Pol}ya trees.
\newblock \emph{Journal of the American Statistical Association}, 97\penalty0
  (460), 2002.

\bibitem[He et~al.(2009)He, Li, Viant, and Yao]{he2009profiling}
Shan He, Xiaoli Li, Mark~R Viant, and Xin Yao.
\newblock Profiling {MS proteomics} data using smoothed non-linear energy
  operator and {Bayes}ian additive regression trees.
\newblock \emph{Proteomics}, 9\penalty0 (17):\penalty0 4176--4191, 2009.

\bibitem[He(1997)]{he1997quantile}
Xuming He.
\newblock Quantile curves without crossing.
\newblock \emph{The American Statistician}, 51\penalty0 (2):\penalty0 186--192,
  1997.

\bibitem[Ishwaran(2015)]{ishwaran2015effect}
Hemant Ishwaran.
\newblock The effect of splitting on random forests.
\newblock \emph{Machine Learning}, 99\penalty0 (1):\penalty0 75--118, 2015.

\bibitem[Kapelner and Bleich(2013)]{kapelner2013bartmachine}
Adam Kapelner and Justin Bleich.
\newblock bartmachine: Machine learning with {Bayes}ian additive regression
  trees.
\newblock \emph{arXiv preprint arXiv:1312.2171}, 2013.

\bibitem[Kindo et~al.(2016)Kindo, Wang, and Pe\~na]{kindo2016mpbart}
Bereket~P. Kindo, Hao Wang, and Edsel~A. Pe\~na.
\newblock Multinomial probit {Bayes}ian additive regression trees.
\newblock \emph{Stat}, pages n/a--n/a, 2016.
\newblock ISSN 2049-1573.
\newblock \doi{10.1002/sta4.110}.
\newblock URL \url{http://dx.doi.org/10.1002/sta4.110}.

\bibitem[Koenker(1994)]{koenker1994confidence}
Roger Koenker.
\newblock Confidence intervals for regression quantiles.
\newblock In \emph{Asymptotic Statistics}, pages 349--359. Springer, 1994.

\bibitem[Koenker and Bassett~Jr(1978)]{koenker1978regression}
Roger Koenker and Gilbert Bassett~Jr.
\newblock Regression quantiles.
\newblock \emph{Econometrica: Journal of the Econometric Society}, pages
  33--50, 1978.

\bibitem[Kordas(2006)]{kordas2006smoothed}
Gregory Kordas.
\newblock Smoothed binary regression quantiles.
\newblock \emph{Journal of Applied Econometrics}, 21\penalty0 (3):\penalty0
  387--407, 2006.

\bibitem[Kottas and Gelfand(2001)]{kottas2001bayesian}
Athanasios Kottas and Alan~E Gelfand.
\newblock {Bayes}ian semiparametric median regression modeling.
\newblock \emph{Journal of the American Statistical Association}, 96\penalty0
  (456):\penalty0 1458--1468, 2001.

\bibitem[Kozumi and Kobayashi(2011)]{kozumi2011gibbs}
Hideo Kozumi and Genya Kobayashi.
\newblock Gibbs sampling methods for {Bayes}ian quantile regression.
\newblock \emph{Journal of Statistical Computation and Simulation}, 81\penalty0
  (11):\penalty0 1565--1578, 2011.

\bibitem[Leisch and Dimitriadou(2010)]{leisch2010mlbench}
Friedrich Leisch and Evgenia Dimitriadou.
\newblock mlbench: Machine learning benchmark problems.
\newblock \emph{R package}, 2:\penalty0 1--1, 2010.

\bibitem[Li et~al.(2010)Li, Xi, and Lin]{li2010bayesian}
Qing Li, Ruibin Xi, and Nan Lin.
\newblock Bayesian regularized quantile regression.
\newblock \emph{Bayesian Analysis}, 5\penalty0 (3):\penalty0 533--556, 09 2010.
\newblock \doi{10.1214/10-BA521}.
\newblock URL \url{http://dx.doi.org/10.1214/10-BA521}.

\bibitem[Liu et~al.(2015)Liu, Traskin, Lorch, George, and
  Small]{liu2015ensemble}
Yang Liu, Mikhail Traskin, Scott~A Lorch, Edward~I George, and Dylan Small.
\newblock Ensemble of trees approaches to risk adjustment for evaluating a
  hospital’s performance.
\newblock \emph{Health Care Management Science}, 18\penalty0 (1):\penalty0
  58--66, 2015.

\bibitem[Marrocu et~al.(2015)Marrocu, Paci, and Zara]{marrocu2015micro}
Emanuela Marrocu, Raffaele Paci, and Andrea Zara.
\newblock Micro-economic determinants of tourist expenditure: A quantile
  regression approach.
\newblock \emph{Tourism Management}, 50:\penalty0 13--30, 2015.

\bibitem[Meinshausen(2006)]{meinshausen2006quantile}
Nicolai Meinshausen.
\newblock Quantile regression forests.
\newblock \emph{The Journal of Machine Learning Research}, 7:\penalty0
  983--999, 2006.

\bibitem[Pedersen(2015)]{pedersen2015predictable}
Thomas~Q Pedersen.
\newblock Predictable return distributions.
\newblock \emph{Journal of Forecasting}, 34\penalty0 (2):\penalty0 114--132,
  2015.

\bibitem[Pratola et~al.(2014)Pratola, Chipman, Gattiker, Higdon, McCulloch, and
  Rust]{pratola2014parallel}
Matthew~T Pratola, Hugh~A Chipman, James~R Gattiker, David~M Higdon, Robert
  McCulloch, and William~N Rust.
\newblock Parallel {Bayes}ian additive regression trees.
\newblock \emph{Journal of Computational and Graphical Statistics}, 23\penalty0
  (3):\penalty0 830--852, 2014.

\bibitem[Qian et~al.(2015)Qian, Yang, and Zou]{qian2015tweedie}
Wei Qian, Yi~Yang, and Hui Zou.
\newblock Tweedie's compound poisson model with grouped elastic net.
\newblock \emph{Journal of Computational and Graphical Statistics}, 0\penalty0
  (ja):\penalty0 0--0, 2015.
\newblock \doi{10.1080/10618600.2015.1005213}.
\newblock URL \url{http://dx.doi.org/10.1080/10618600.2015.1005213}.

\bibitem[Reich et~al.(2010)Reich, Bondell, and Wang]{reich2010flexible}
Brian~J Reich, Howard~D Bondell, and Huixia~J Wang.
\newblock Flexible {Bayes}ian quantile regression for independent and clustered
  data.
\newblock \emph{Biostatistics}, 11\penalty0 (2):\penalty0 337--352, 2010.

\bibitem[Rudnicki et~al.(2015)Rudnicki, Wrzesie{\'n}, and
  Paja]{rudnicki2015all}
Witold~R Rudnicki, Mariusz Wrzesie{\'n}, and Wies{\l}aw Paja.
\newblock All relevant feature selection methods and applications.
\newblock In \emph{Feature Selection for Data and Pattern Recognition}, pages
  11--28. Springer, 2015.

\bibitem[Sriram et~al.(2013)Sriram, Ramamoorthi, and
  Ghosh]{sriram2013posterior}
Karthik Sriram, RV~Ramamoorthi, and Pulak Ghosh.
\newblock Posterior consistency of {Bayes}ian quantile regression based on the
  misspecified asymmetric {Lapl}ace density.
\newblock \emph{{Bayes}ian Analysis}, 8\penalty0 (2):\penalty0 479--504, 2013.

\bibitem[Taddy and Kottas(2012)]{taddy2012bayesian}
Matthew~A Taddy and Athanasios Kottas.
\newblock A {Bayes}ian nonparametric approach to inference for quantile
  regression.
\newblock \emph{Journal of Business \& Economic Statistics}, 2012.

\bibitem[Tsai(2012)]{tsai2012relationship}
I-Chun Tsai.
\newblock The relationship between stock price index and exchange rate in asian
  markets: A quantile regression approach.
\newblock \emph{Journal of International Financial Markets, Institutions and
  Money}, 22\penalty0 (3):\penalty0 609 -- 621, 2012.
\newblock ISSN 1042-4431.
\newblock \doi{http://dx.doi.org/10.1016/j.intfin.2012.04.005}.
\newblock URL
  \url{http://www.sciencedirect.com/science/article/pii/S1042443112000297}.

\bibitem[Wu et~al.(2010)Wu, Huang, and Pan]{wu2010novel}
Jiansheng Wu, Liangyong Huang, and Xiaoming Pan.
\newblock A novel {Bayes}ian additive regression trees ensemble model based on
  linear regression and nonlinear regression for torrential rain forecasting.
\newblock In \emph{Computational Science and Optimization (CSO), 2010 Third
  International Joint Conference on}, volume~2, pages 466--470. IEEE, 2010.

\bibitem[Yu and Moyeed(2001)]{yu2001bayesian}
Keming Yu and Rana~A Moyeed.
\newblock {Bayes}ian quantile regression.
\newblock \emph{Statistics \& Probability Letters}, 54\penalty0 (4):\penalty0
  437--447, 2001.

\bibitem[Zhang and H{\"a}rdle(2010)]{ZhaEtAl2010}
Junni~L Zhang and Wolfgang~K H{\"a}rdle.
\newblock The {Bayes}ian additive classification tree applied to credit risk
  modelling.
\newblock \emph{Computational Statistics \& Data Analysis}, 54\penalty0
  (5):\penalty0 1197--1205, 2010.

\bibitem[Zou and Yuan(2008)]{zou2008composite}
Hui Zou and Ming Yuan.
\newblock Composite quantile regression and the oracle model selection theory.
\newblock \emph{The Annals of Statistics}, pages 1108--1126, 2008.

\end{thebibliography}
\bibliographystyle{plainnat}
\end{document}